\documentclass[10pt, a4paper]{article}

\usepackage{lrec-coling2024} 
\usepackage{booktabs}
\usepackage{multirow}
\usepackage{natbib}
\usepackage{multibib}
\usepackage{fontawesome}

\makeatletter
\def\@mb@citenamelist{cite,citep,citet,citealp,citealt,citepalias,citetalias}
\makeatother
\newcites{languageresource}{~}

\usepackage{graphicx}
\usepackage{tabularx}
\usepackage{soul}
\usepackage{titlesec}
\titleformat{\section}{\normalfont\large\bfseries\center}{\thesection.}{1em}{}
\titleformat{\subsection}{\normalfont\SmallTitleFont\bfseries\raggedright}{\thesubsection.}{1em}{}
\titleformat{\subsubsection}{\normalfont\normalsize\bfseries\raggedright}{\thesubsubsection.}{1em}{}
\renewcommand\thesection{\arabic{section}}
\renewcommand\thesubsection{\thesection.\arabic{subsection}}
\renewcommand\thesubsubsection{\thesubsection.\arabic{subsubsection}}

\usepackage{xcolor}
\usepackage{hyperref}
 \definecolor{darkblue}{rgb}{0, 0, 0.5}
  \hypersetup{colorlinks=true, citecolor=darkblue, linkcolor=darkblue, urlcolor=darkblue}

\usepackage{xstring}

\usepackage{color}

\title{Self-Explanation Prompting Improves Dialogue Understanding \\ in Large Language Models}
\name{
    \begin{tabular}{c}
        Haoyu Gao$^{123 \dagger}$, Ting-En Lin$^{2}$, Hangyu Li$^{2}$, Min Yang$^{3*}$, \\ Yuchuan Wu$^{2}$, Wentao Ma$^{2}$, Yongbin Li$^{2*}$
    \end{tabular}
    \thanks{$^{*}$ Corresponding authors.}
    \thanks{$\dagger$ Work done while interning at Alibaba.}
}
\address{
    $^{1}$ University of Science and Technology of China \quad  $^{2}$ Alibaba Group \\
    $^{3}$Shenzhen Institute of Advanced Technology, Chinese Academy of Sciences\\
    \texttt{\{hy.gao, min.yang\}@siat.ac.cn} \\
    \texttt{shuide.lyb@alibaba-inc.com}
}

\abstract{
Task-oriented dialogue (TOD) systems facilitate users in executing various activities via multi-turn dialogues, but Large Language Models (LLMs) often struggle to comprehend these intricate contexts. In this study, we propose a novel "Self-Explanation" prompting strategy to enhance the comprehension abilities of LLMs in multi-turn dialogues. This task-agnostic approach requires the model to analyze each dialogue utterance before task execution, thereby improving performance across various dialogue-centric tasks. Experimental results from six benchmark datasets confirm that our method consistently outperforms other zero-shot prompts and matches or exceeds the efficacy of few-shot prompts, demonstrating its potential as a powerful tool in enhancing LLMs' comprehension in complex dialogue tasks.
 \\ \newline \Keywords{large language model, prompting, dialog understanding}}

\begin{document}

\maketitleabstract

\section{Introduction}
Recent advancements in large language models (LLMs) have achieved great success in various NLP tasks. \cite{gpt3, llama2, chowdhery2022palm}. However, the vast model parameters pose challenges in downstream fine-tuning. To circumvent these challenges, diverse zero-shot prompting strategies have been researched to enhance LLM performance \cite{gpt3, liu2021makes, sorensen2022information}. \textit{In-context learning} emerges as a viable alternative to fine-tuning, leveraging examples to augment language processing abilities. To elicit the reasoning ability of LLMs, Chain-of-Thought has been seamlessly integrated within the prompting framework, showing remarkable performance in tasks requiring complex reasoning \cite{cot, cot_arithmetic, cot_comonsense}. Stemming from CoT prompting, numerous studies have delved into refining CoT via prompt design modifications \cite{li2022advance, fu2022complexity, zhang2022automatic} and optimizing reasoning paths \cite{self, wang2022rationale, zelikman2022star}. In contrast, to reduce dependency on human demonstrations, the Zero-shot CoT \cite{zero-shot_cot} employs the post-append instruction, 'Let's think step by step,' urging Large Language Models (LLMs) to derive the stages of reasoning sequentially and automatically.



\begin{figure}[t]
    \centering
    \includegraphics[width=\linewidth]{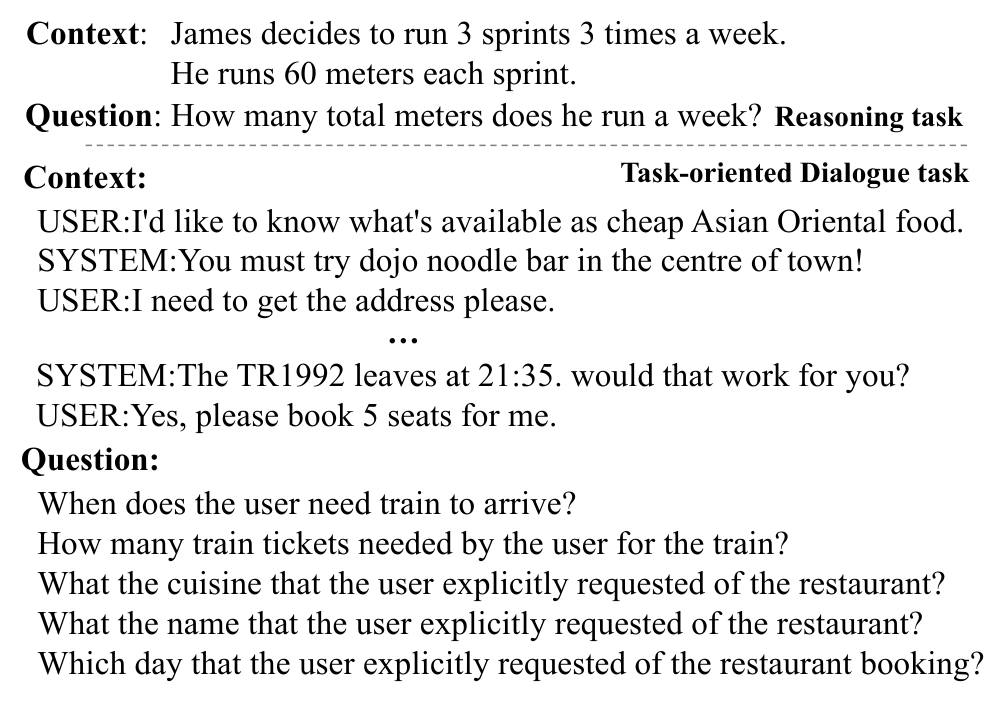}
    \caption{The input example for the reasoning task and the task-oriented dialogue is structured into two components: Context and Question.}
    \label{running_example}
\end{figure}

\begin{table*}[t]
\centering
\caption{Comparative analysis of reasoning and dialogue understanding tasks, highlighting the distinctive application of the proposed Self-Explanation method.}
\begin{tabular}{ccccc|cc}
    \toprule
    \textbf{Task} &
      \textbf{Dataset} &
      \textbf{\begin{tabular}[c]{@{}c@{}}Avg. \\ \#Tokens\end{tabular}} &
      \textbf{Context} &
      \textbf{\begin{tabular}[c]{@{}c@{}}Answer \\ Search Space\end{tabular}} &
      \textbf{\begin{tabular}[c]{@{}c@{}}Prompting\\ Method\end{tabular}} &
      \textbf{Focus on} \\
    \midrule
    Reasoning &
      \begin{tabular}[c]{@{}c@{}}MultiArith\\ GSM8K\end{tabular} &
      \begin{tabular}[c]{@{}c@{}}16.6\\ 33.6\end{tabular} &
      Short &
      Internal &
      \begin{tabular}[c]{@{}c@{}}Chain-of-Thought\\ Plan-and-Solve\end{tabular} &
      \begin{tabular}[c]{@{}c@{}}Reasoning\\ Step\end{tabular} \\
    \midrule
    \begin{tabular}[c]{@{}c@{}}Dialog\\ Understanding\end{tabular} &
      \begin{tabular}[c]{@{}c@{}}SGD\\ MultiWOZ\end{tabular} &
      \begin{tabular}[c]{@{}c@{}}940.9\\ 1229.7\end{tabular} &
      Long &
      External &
      Self-Explanation &
      Context \\
    \bottomrule
\end{tabular}
\label{token_num}
\end{table*}

Despite the effectiveness of CoT prompting, most existing prompting methods focus on eliciting the reasoning ability inherent in large language models. However, these techniques might fall short when applied to tasks that require contextual comprehension rather than reasoning steps. Specifically, dialogue-based tasks \cite{lin2022duplex, hu2022unimse, li2023unisa} serve as typical examples that require strong comprehension ability rather than reasoning ability. The task-oriented dialogue (TOD) \cite{he2022space, he2022unified, he2022galaxy} is one of the most representative tasks that facilitates users in executing various activities, including but not limited to hotel and restaurant reservations, by engaging in multi-turn dialogues \cite{gao2023unsupervised, qian2023empathetic, yu2023speech}. An illustrative example of both the reasoning task and the TOD can be seen in Figure \ref{running_example}. Contrary to the reasoning task, which typically consists of concise context, the TOD mostly involves multi-turn dialogues with long contexts. Not only do these tasks differ in terms of context length, but they also exhibit variations across numerous other dimensions. For instance, as delineated in Table \ref{token_num}, the reasoning task predominantly emphasizes intricate problem-solving steps that entail extensive computations and conversions. This underscores the model's inherent ability to reason. Consequently, the scope of searching for an answer predominantly resides within the model.



However, when performing dialogue-based tasks, success depends on a strong understanding of the context in continuous conversational exchanges rather than complex reasoning. TOD tasks mainly obtain information directly from the existing context, making the search space for answers strongly related to external contexts. The different emphases of the two tasks resulted in the underperformance of CoT prompts in dialogue contexts. Judging from the results of existing evaluation studies \cite{gasic, 24, multitask}, the current LLMs with unoptimized prompting perform significantly worse than specialized small models on some dialogue-based tasks. \citet{text_sql} have reformulated the dialogue state tracking task into a few-shot text-to-SQL paradigm, utilizing the robust code capabilities of Codex. While this represents an intriguing approach for training dialogue exemplars, the text-to-SQL may not be universally applicable, particularly in procedural TOD tasks such as next-action prediction. Additionally, the example retriever needs to be retrained for each new dataset, which imposes limitations on this approach.

To address the above issues, we explore several ways to enhance the comprehension capabilities of LLMs by mimicking the way humans solve conversational problems \cite{chi1989self}. We introduce the "Self-Explanation" prompt strategy, requiring the model to explain every utterance in the dialogue first and then complete the task based on the generated explanation. Despite its simplicity, the proposed method enhances the performance of contextual comprehension of LLMs in various dialogue-centric tasks. 
More importantly, our prompt is task-agnostic and can be easily applied to a variety of problems involving multi-turn dialogue. 
We evaluate the proposed method across six dialogue-centric datasets. The results show that our prompt consistently surpasses other zero-shot prompts and is on par with or surpasses few-shot prompts. In summary, our contributions include:


\begin{itemize}
\item We conduct a comprehensive comparison between reasoning tasks and dialogue understanding tasks, identifying the limitations of current prompting methods.
\item We propose a simple yet effective prompting strategy, Self-Explanation, that significantly enhances the dialogue comprehension capacities of large language models.
\item Extensive experiments on six dialogue-based datasets have demonstrated that the proposed method surpasses existing prompting approaches in performance.
\end{itemize}

\begin{figure*}[t]
    \centering
    \includegraphics[width=\linewidth]{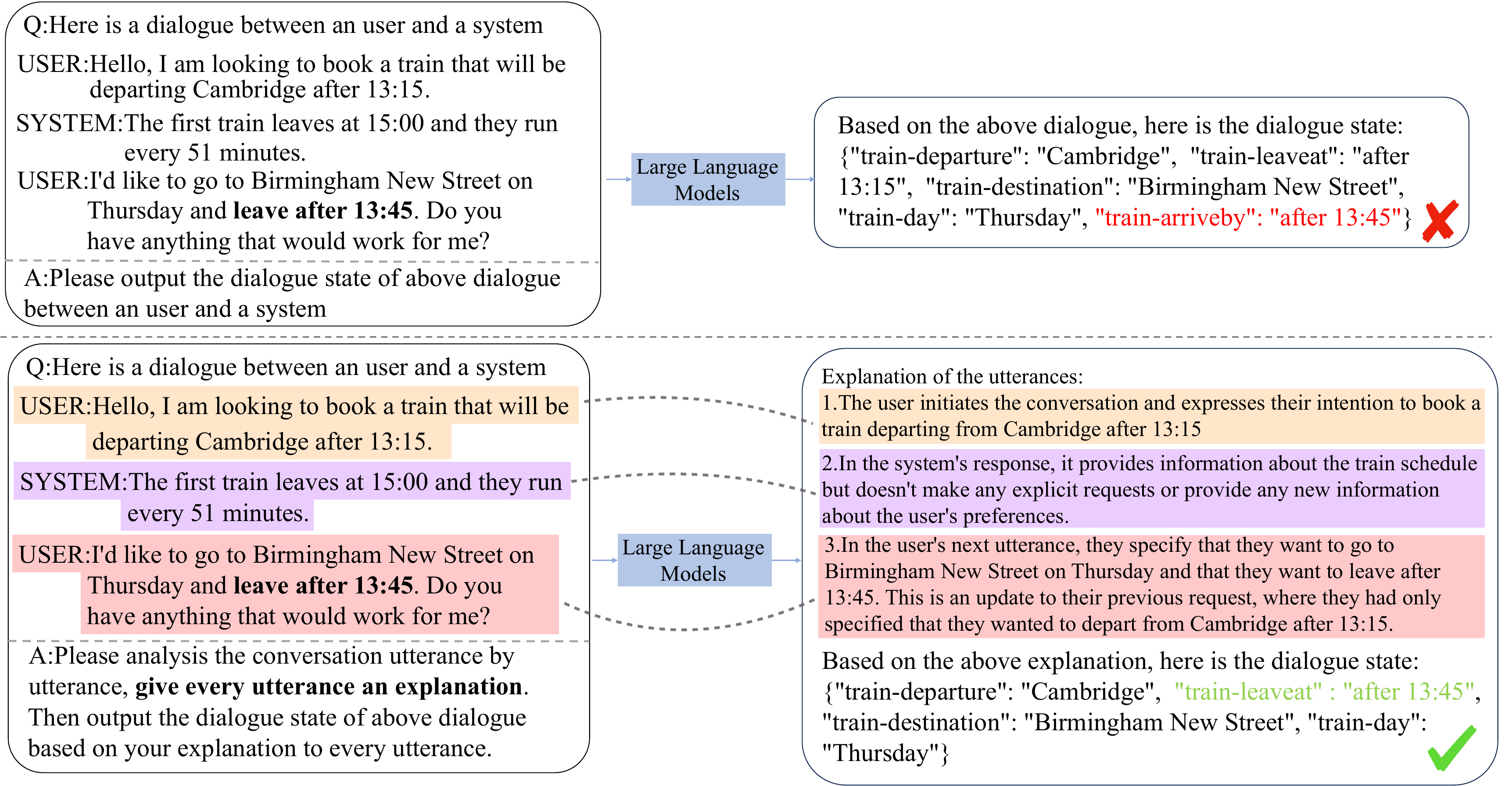}
    \caption{Example inputs and outputs of GPT-3 with No explanation ahead (upper) and Explain before answer (lower). Explanation greatly improves the understanding of the dialogue.}
    \label{fig:model}
\end{figure*}

\section{Method}
\subsection{Formalization}

The problem can be divided into two components: the context, denoted as $\mathcal{C}$, and the question, represented by $\mathcal{Q}$. The context, $\mathcal{C}$, provides a descriptive framework that outlines the problem setting and background. For reasoning tasks, this context delineates a specific situation. An example of this can be observed in Figure \ref{running_example}, where $\mathcal{C}$ contains the activities of James. Meanwhile, in the context of TOD tasks, $\mathcal{C}$ typically captures a multi-turn dialogue between two interlocutors.

Contrastingly, the question component, $\mathcal{Q}$, zeroes in on a specific inquiry related to $\mathcal{C}$. In the realm of reasoning tasks, $\mathcal{Q}$ typically solicits a value derived from multi-step computations. This implies that the solution isn't readily available within $\mathcal{C}$. To illustrate, refer to Figure \ref{running_example} where $\mathcal{Q}$ probes for the aggregate distance James covers in a week. Addressing this necessitates discerning the frequency of James' sprints per week and the distance of each sprint. Subsequent multiplication of these two quantities yields the desired result.

On the other hand, in a TOD task, the nature of $\mathcal{Q}$ is more straightforward, often inquiring about the existence of specific information. Using Figure \ref{running_example} as a reference, the question might pertain to the scheduled departure time of a reserved train or the type of cuisine a user seeks. Responses to these types of inquiries are readily extractable from $\mathcal{C}$, obviating the need for additional computation.

\subsection{Self-Explanation}

Humans often find it challenging to respond to questions grounded in extensive new information.
One strategy that has been empirically shown to enhance comprehension of new material is self-explanation. The concept of self-explanation, originating from psychological research \cite{chi1989self}, involves learners generating explanations for themselves while processing unfamiliar content. Notably, this study demonstrated that learners engaging in self-explanation were better able to grasp core concepts and principles than their counterparts who did not employ this strategy.

Drawing inspiration from human cognitive processes and this psychological paradigm, we introduce the Self-Explanation prompting method, a zero-shot prompting technique designed to enhance multi-turn dialogue comprehension. Within the process, models initially provide explanations for each utterance in a multi-turn dialogue. Subsequently, these models execute the specified task, relying on their previously generated explanations. In the process of articulation, the large language models (LLMs) have the capacity to transform low-level natural language inputs into more abstract, high-level constructs, such as the intent or action of the speaker.

The framework is structured without the need for demonstration examples. Following the problem formalization in section 2.1, we organise the inputs using the template "$\mathcal{C}$:[C]. $\mathcal{Q}$:[Q]. $\mathcal{A}$:[A]", wherein [C] and [Q] represents the input slot designated for the context and question, respectively. As for the last part, [A] is populated by manually-curated instructions prompting the model to elucidate. Central to our method is the instruction: \textit{"Provide explanations for each utterance and then respond based on these explanations."} For the decoding strategy, we opt for the straightforward greedy decoding method, though beam search decoding could be employed to produce a broader range of explanations.

\begin{table*}[t]
\centering
\caption{Comparing the performance of zero-shot CoT and Self-Explanation methods on six dialogue datasets using gpt-3.5-turbo.}
\label{main_results}
\begin{tabular}{ccccc|c|c}
\toprule
                              & \multicolumn{4}{c}{TOD}                                           & \multicolumn{1}{c}{ERC}            & \multicolumn{1}{c}{RS}             \\ \cmidrule{2-7}
    Method                    & MutliWoz 2.1       & STARv2         & SGD            & SpokenWoz      & MELD           & MuTual         \\
\midrule
    Vanilla                   & 35.93          & 51.88          & 18.96          & 13.75          & 59.14          & 68.97          \\
    Vanilla + 4-shots      & 41.60           & 52.93          & 17.34          & 14.13          & 55.09          & \textbf{72.51} \\
    Chain-of-Thought          & 27.64          & 51.85          & 19.69          & 13.26          & 61.48          & 70.61          \\
    Plan-and-Solve            & 39.19          & 56.74           & 21.11          & 14.50           & 58.38          & 69.77          \\
\midrule
    \textbf{Self-Explanation} & \textbf{44.44} & \textbf{63.66} & \textbf{21.81} & \textbf{14.89} & \textbf{61.71} & 71.58          \\    
    \quad {+ GPT-4}   & 50.97 & 70.27 & 25.75 & 25.94 & 63.51 &  91.87        \\    
\bottomrule
\end{tabular}
\end{table*}


\section{Experiments}
\subsection{Experimental Setup}
\subsubsection{Datasets and task}
We evaluate our self-explanation on six datasets from three categories of dialogue understanding tasks: Task-oriented dialogue (TOD), Emotion Recognition in Conversations (ERC) task and Response Selection (RS) task.
For TOD task, the datasets can be divided into two types based on the dialogue schema: Procedural and Declarative \cite{schema_cate}. A dialogue schema in the context of task-oriented dialogue is a structured representation of the conversation flow or some key entities (also known as ’slots’) that need to be captured.
The Procedural schema, derived from the STAR dataset \cite{star}, represents a dialogue domain as a directed graph similar to a flowchart. It consists of nodes representing user utterances, system responses, or backend service calls. The main task of the procedural schema is to strictly follow the task flow. 
For Procedural schema, we choose STARv2 \cite{starv2} dataset.

\textbf{STARv2} dataset, which is an upgraded version of STAR \cite{star} with new ground truth belief state and new natural language action descriptions. STAR is a schema-guided task-oriented dialogue dataset consisting of 24 tasks across 13 domains. We evaluate the next action prediction task, which is to predict the next system action conditioned on the dialogue history and take the weighted F-1 score as the metric.

The Declarative format, based on the Schema-Guided Dialogue (SGD) dataset \cite{sgd} and MultiWOZ dataset \cite{multiwoz}, aims to capture the slots defined in dataset ontology. For the declarative format schema, we select MultiWOZ 2.1, SGD, and SpokenWOZ \cite{spokenwoz} dataset and evaluate the dialogue state tracking task, using Joint Goal Accuracy (JGA) as the metric.

\textbf{MultiWOZ2.1} is a fully-labeled collection of human-human written conversations spanning multiple domains and topics. It contains 7 domains, 35 slots, and over 10k dialogues. 

\textbf{SGD} is another declarative format dataset containing over 16k multi-domain conversations spanning 16 domains with more slots and possible values compared to MultiWOZ. 

\textbf{SpokenWOZ} is a new multi-modal spoken TOD dataset containing 8 domains, 5.7k dialogues, and 35 slots. It introduces the unique challenges in spoken conversation. 

Besides the task-oriented dialogue, we also choose two datasets: \textbf{MELD} \cite{meld} and \textbf{MuTual} \cite{mutual} from the Emotion Recognition in Conversations (ERC) task and response selection task, respectively. MELD contains over 10k utterances from the TV series Friends, and each utterance is annotated with emotion and sentiment labels. MuTual consists of 8k manually annotated dialogues based on Chinese students English listening comprehension exams.

\begin{table*}[t]
\caption{The effect of different trigger sentences measured on MultiWOZ with gpt3.5-turbo.}
\label{ablation_result}
\begin{tabular}{llc}
\toprule
Method & Prompt & MultiWOZ 2.1(JGA) \\
\midrule
Vanilla & Answer the questions based on the above dialogue  & 35.93  \\
\midrule
Understand & \begin{tabular}[c]{@{}l@{}}Before you answer, \textbf{first understand the dialogue}, then answer \\ the questions based on your understanding and original dialogue\end{tabular} & 36.52  \\
\midrule
Summary  & \begin{tabular}[c]{@{}l@{}}Before you answer, \textbf{first summarize the dialogue}, then answer \\ the questions based on your summary and original dialogue\end{tabular} & 40.98 \\
\midrule
Explanation & \begin{tabular}[c]{@{}l@{}}Before you answer, first analyze the dialogue utterance  \\ by utterance, \textbf{give every utterance an explanation}. \\ Then answer the questions based on your explanation\end{tabular} & 44.44 \\
\bottomrule
\end{tabular}
\end{table*}

\subsubsection{Baselines}
We compare our proposed zero-shot Explanation with two types of prompt baselines: Zero-shot baselines and Few-shot. For zero-shot baselines, we include zero-shot-CoT \cite{zero-shot_cot} and Plan-and-Solve Prompting \cite{plan-and-solve}. The former appends “Let’s think step by step” to the prompt. The latter extends the zero-shot-CoT with plan ahead, and then carry out the plan. 
Besides the zero-shot baselines, we also evaluate the In-Context learning prompt performance on TOD task. Considering the sample of TOD task consists of a multi-turn dialogue and the slot list, we only use 4 examples as for not exceed the context window size. As for example selection, we randomly selected 4 examples with the same domain as the test sample.

\subsection{Main Results}
Table \ref{main_results} presents the performance of our method compared to baseline approaches across six distinct datasets. In the zero-shot scenario, our technique consistently surpasses the baselines on all evaluation datasets, irrespective of their differences. While CoT prompting does not enhance performance on TOD tasks, our method notably excels by an impressive 12\% margin on the STARv2 dataset.

This significant improvement underscores the effectiveness of self-explanation prompting. The task format aligns well with this prompting approach, leading to detailed sentence-by-sentence explanations. These explanations play a pivotal role in comprehending the dialogue flow and adhering to the given schema. The enhanced performances on MultiWOZ, SGD, and SpokenWOZ further affirm that the dialogue state tracking task greatly benefits from self-explanation prompts. By providing explanations for each utterance, the likelihood of overlooking dialogue states is diminished. In addition to the task-oriented dialogue tasks, we assessed the impact of self-explanation prompting on both the ERC and RS tasks. However, the gains here were relatively modest in comparison to the TOD tasks. Given that our explanations are rooted in semantic interpretations, they may not be as beneficial for tasks centered on emotion recognition.

Compared to the few-shot baseline, our zero-shot prompting either outperforms or matches performance across all six datasets. This underscores the argument that a comprehensive understanding of dialogue is more critical than merely having a set of examples. The efficacy of in-context learning is largely attributed to its input-label pairing formats, its access to the label space, and the modification of the input distribution. For tasks within the domain of TOD, the input usually consists of multi-turn dialogues encompassing various topics, necessitating a profound understanding of the dialogue's entirety. The intricate nature of TOD tasks demands a high level of comprehension, which mere exposure to a few examples fails to deliver.


\subsection{Analysis}
\subsubsection{Effect of Explanation}

To assess the impact of self-explanation on dialogue comprehension, we carried out a comparative study using the MultiWOZ dataset, testing four distinct prompting methods. The results of these tests can be found in \ref{ablation_result}. 

In the \textbf{Vanilla} method, no additional instruction is given before the model provides its response.
In the \textbf{Understand} method, the model is simply prompted with "Understand the dialogue first" prior to answering. However, there's no specified format for the intermediate comprehension.
With the \textbf{Summary} method, the model is prompted to first summarize the dialogue. It then bases its answer on both the summary and the original dialogue.

Our observations revealed that when comparing the self-explanation method with Vanilla, there was a notable decline in performance. This suggests that pre-processing or understanding the dialogue is essential for optimal performance. Merely prompting the model to understand the dialogue without detailed instruction also resulted in reduced performance. This demonstrates the importance of precise comprehension guidelines. Without them, LLMs tend to produce explanations for their answers as opposed to comprehending the dialogue. Providing detailed comprehension instructions is less ambiguous than allowing the model to self-navigate. The Summary method explicitly directs the model to use the summary as a means of comprehension, subsequently answering based on that summary. This approach enhanced performance by approximately 5\% JGA in comparison to the Vanilla method. However, summarizing is a broad-strokes approach and might overlook finer details essential for the TOD task. 

Drawing from psychological research, specifically \cite{chi1989self}, it's evident that not all explanations confer the same benefits. Factors like content, quality, and depth of explanations are paramount. Our refined method, the \textbf{self-explanation} prompting, instructs the model to generate sequential explanations, promoting deeper dialogue understanding.


\begin{table*}[t]
\caption{The dialogue content and answer predicted of three typical error type in MultiWOZ dataset where Self-Explanation get the correct answer and Vanilla get the incorrect answer.}
\label{case_study}
\small
\begin{tabular}{llcc}
\toprule
Error Type  & Dialogue & Explanation  & Vanilla  \\
\midrule
Time involved & \faUser \medspace I need to \textbf{get to} Michaelhouse cafe \textbf{by 12:45}.& \begin{tabular}[c]{@{}c@{}}taxi-arriveby: \\ 12:45\end{tabular}  & \begin{tabular}[c]{@{}c@{}}\textcolor{red}{taxi-leaveat:}\\ 12:45\end{tabular} \\
\midrule
Missing info.  & \begin{tabular}[c]{@{}l@{}}\faUser \medspace I am \textbf{leaving Cambridge} at 12:00 on Sunday, \\ can you please tell me the travel time on that ride?\end{tabular} & \begin{tabular}[c]{@{}c@{}}train-departure:\\ cambridge\end{tabular} & \begin{tabular}[c]{@{}c@{}}train-departure:\\ \textcolor{red}{None}\end{tabular} \\
\midrule
\begin{tabular}[c]{@{}l@{}}Task unclear \\ understand\end{tabular} & \begin{tabular}[c]{@{}l@{}}\faUser \medspace Please help me find the \textbf{attraction downing college}.\\ \faLaptop \medspace Yes, it's on Regent Street \textbf{in the centre of town}. \\  Would you like the phone number?\end{tabular} & \begin{tabular}[c]{@{}c@{}}attraction-name: \\ downing college\end{tabular} & \begin{tabular}[c]{@{}c@{}}attraction-name: \\ downing college \\ \textcolor{red}{ attraction-area:}\\ \textcolor{red}{centre} \end{tabular} \\
\bottomrule
\end{tabular}
\end{table*}

\subsubsection{Case Study}
To have a straightforward understanding of how explanation affects task completion, We manually checked all the cases of MultiWOZ dataset that were correctly answered by self-explanation but incorrectly answered by Vanilla and picked several typical errors. As shown in Table \ref{case_study}

Generally, there are three main types of errors: time involved, missing information, and unclear task understanding. 
For the first error type, the model usually gets confused with departure time and arrival time. In the case of time-involved errors, the user needs a taxi that arrives by 12:45. While the model output of the vanilla prompting assigns 12:45 to the time of taxi departure.

The second error type, missing information error, mostly happens in the dialogue, which has a high number of turns. The large amount of dialogue information may distract the model from correctly capturing all the information needed to complete the task. As the case of this error shows, the user expresses the place, time, and date of departure in one sentence. The model output of vanilla prompting misses the place of departure, while the output of self-explanation prompting correctly captures all the information about the user request.

The last error type is a task-specific error. In the dialogue state track task, the dialogue state should include the information that the user requested and exclude the system provides.
In the case of the final type of error, the user explicitly requests an attraction called Downing College, and the system provides some relevant information about this attraction. The model output of self-explanation prompting correctly distinguishes the information the user requested and the system provided. While the model output of Vanilla prompting mistakenly includes the system information in the dialogue state.

\subsubsection{Connection with CoT Prompting}
We have explored self-explanation prompting as a simple way to enhance the understanding of multi-turn dialogue in large language models. In this section, we'll connect the dots between self-explanation prompting and CoT prompting.

From a macro perspective, OpenAI's documentation indicates that giving models a moment to "think" is beneficial. Analogous to human cognition, hastily jumping to conclusions often leads to mistakes. CoT prompting, which requires a systematic rationale before presenting an answer, effectively grants models this "thinking" time. Similarly, our self-explanation prompting offers models a moment of reflection, but it steers them to interpret the intricate context, 
$\mathcal{C}$, as opposed to breaking down the answer's rationale.

From a micro perspective, CoT prompting guides the model toward a solution by narrowing the scope of potential answers. In tasks requiring reasoning, the solution isn't straightforwardly derived from the context 
$\mathcal{C}$. The response involves extensive calculations and transformations, heavily drawing on the model's innate reasoning faculties. This suggests the solution space is largely tethered to the model's capabilities. The logical progression elicited by CoT prompting either constrains or directs this solution space.

Conversely, in the TOD task, the query 
$\mathcal{Q}$ typically seeks details readily found in 
$\mathcal{C}$. Unlike reasoning assignments, these questions don't demand intricate computations. As such, the solution space primarily lies within 
$\mathcal{C}$. The enhanced dialogue comprehension, courtesy of our self-explanation prompting, offers an alternative approach to narrowing down this solution space.

 
\section{Related work}

\textbf{Prompting Methods:} The exploration of prompting methods for machine learning models has been vast. One of the conventional methods is in-context learning (ICL), as highlighted by GPT-3\cite{gpt3}. In ICL, multiple demonstrations are provided before a test sample, and the model's performance significantly hinges on these demonstrations \cite{zhao2021calibrate, lu2021fantastically}.

Some researchers, such as those in \citet{liu2021makes}, endeavor to retrieve examples semantically similar to a test query sample, utilizing metrics like the L2 distance or cosine-similarity distance derived from sentence embeddings. In addition to these distance metrics, the concept of mutual information emerges as a potent example selection criterion \cite{sorensen2022information}. Here, the goal is to select a template that optimizes the mutual information between the input and the model's output.
Taking this further, several studies, such as \citep{rubin2021learning}, have shifted towards a supervised approach, training models to pick the most relevant demonstrations from a pool of candidates.

\textbf{Reasoning Strategies:} Beyond merely selecting examples, their arrangement or ordering can significantly influence a model's performance. Enter the Chain-of-Thought (CoT) strategy \cite{cot}, a pioneering prompting approach designed to enhance the performance of large language models (LLMs) on intricate reasoning tasks. Unlike ICL, which relies on prepending input-output pairs, CoT integrates a sequence of intermediate reasoning steps into the demonstration, thereby amplifying the reasoning capabilities of LLMs.

Recognizing the importance of diverse reasoning paths, the self-consistency strategy \cite{self} was introduced. It first creates multiple reasoning paths rather than just the most likely one and subsequently selects the most coherent answer by considering all the generated paths. Further automation in this domain is achieved with zero-shot CoT \cite{zero-shot_cot}. Instead of relying on human-annotated reasoning sequences, this method induces the model to generate reasoning steps by simply prompting it to 'think step by step'.


\section{Conclusion}
In this paper, we find that CoT prompting is suboptimal for multi-turn dialogue tasks that require strong comprehension abilities. To enhance the comprehension of LLM, we propose a new zero-shot prompting strategy called self-explanation prompting, which guides the LLM to first understand the multi-turn dialogue by explaining every utterance and then completing the task based on dialogue with its explanation. Extensive experiments show that explanation prompting can boost the LLMs contextual understanding of multi-turn dialogue and significantly outperform or perform on par with the previous zero-shot and few-shot baselines.


\small
\bibliographystyle{lrec-coling2024-natbib}
\bibliography{lrec-coling2024}

\begin{thebibliography}{0}
\expandafter\ifx\csname natexlab\endcsname\relax\def\natexlab#1{#1}\fi

\end{thebibliography}


\begin{thebibliography}{41}
\expandafter\ifx\csname natexlab\endcsname\relax\def\natexlab#1{#1}\fi

\bibitem[{Bang et~al.(2023)Bang, Cahyawijaya, Lee, Dai, Su, Wilie, Lovenia, Ji,
  Yu, Chung et~al.}]{multitask}
Yejin Bang, Samuel Cahyawijaya, Nayeon Lee, Wenliang Dai, Dan Su, Bryan Wilie,
  Holy Lovenia, Ziwei Ji, Tiezheng Yu, Willy Chung, et~al. 2023.
\newblock A multitask, multilingual, multimodal evaluation of chatgpt on
  reasoning, hallucination, and interactivity.
\newblock \emph{arXiv preprint arXiv:2302.04023}.

\bibitem[{Brown et~al.(2020)Brown, Mann, Ryder, Subbiah, Kaplan, Dhariwal,
  Neelakantan, Shyam, Sastry, Askell et~al.}]{gpt3}
Tom Brown, Benjamin Mann, Nick Ryder, Melanie Subbiah, Jared~D Kaplan, Prafulla
  Dhariwal, Arvind Neelakantan, Pranav Shyam, Girish Sastry, Amanda Askell,
  et~al. 2020.
\newblock Language models are few-shot learners.
\newblock \emph{Advances in neural information processing systems},
  33:1877--1901.

\bibitem[{Budzianowski et~al.(2018)Budzianowski, Wen, Tseng, Casanueva, Ultes,
  Ramadan, and Ga{\v{s}}i{\'c}}]{multiwoz}
Pawe{\l} Budzianowski, Tsung-Hsien Wen, Bo-Hsiang Tseng, Inigo Casanueva,
  Stefan Ultes, Osman Ramadan, and Milica Ga{\v{s}}i{\'c}. 2018.
\newblock Multiwoz--a large-scale multi-domain wizard-of-oz dataset for
  task-oriented dialogue modelling.
\newblock \emph{arXiv preprint arXiv:1810.00278}.

\bibitem[{Chi et~al.(1989)Chi, Bassok, Lewis, Reimann, and
  Glaser}]{chi1989self}
Michelene~TH Chi, Miriam Bassok, Matthew~W Lewis, Peter Reimann, and Robert
  Glaser. 1989.
\newblock Self-explanations: How students study and use examples in learning to
  solve problems.
\newblock \emph{Cognitive science}, 13(2):145--182.

\bibitem[{Chowdhery et~al.(2022)Chowdhery, Narang, Devlin, Bosma, Mishra,
  Roberts, Barham, Chung, Sutton, Gehrmann et~al.}]{chowdhery2022palm}
Aakanksha Chowdhery, Sharan Narang, Jacob Devlin, Maarten Bosma, Gaurav Mishra,
  Adam Roberts, Paul Barham, Hyung~Won Chung, Charles Sutton, Sebastian
  Gehrmann, et~al. 2022.
\newblock Palm: Scaling language modeling with pathways.
\newblock \emph{arXiv preprint arXiv:2204.02311}.

\bibitem[{Cui et~al.(2020)Cui, Wu, Liu, Zhang, and Zhou}]{mutual}
Leyang Cui, Yu~Wu, Shujie Liu, Yue Zhang, and Ming Zhou. 2020.
\newblock Mutual: A dataset for multi-turn dialogue reasoning.
\newblock \emph{arXiv preprint arXiv:2004.04494}.

\bibitem[{Fu et~al.(2022)Fu, Peng, Sabharwal, Clark, and
  Khot}]{fu2022complexity}
Yao Fu, Hao Peng, Ashish Sabharwal, Peter Clark, and Tushar Khot. 2022.
\newblock Complexity-based prompting for multi-step reasoning.
\newblock \emph{arXiv preprint arXiv:2210.00720}.

\bibitem[{Gao et~al.(2023)Gao, Wang, Lin, Wu, Yang, Huang, and
  Li}]{gao2023unsupervised}
Haoyu Gao, Rui Wang, Ting-En Lin, Yuchuan Wu, Min Yang, Fei Huang, and Yongbin
  Li. 2023.
\newblock Unsupervised dialogue topic segmentation with topic-aware contrastive
  learning.
\newblock In \emph{Proceedings of the 46th International ACM SIGIR Conference
  on Research and Development in Information Retrieval}, pages 2481--2485.

\bibitem[{He et~al.(2022{\natexlab{a}})He, Dai, Hui, Yang, Cao, Dong, Huang,
  Si, and Li}]{he2022space}
Wanwei He, Yinpei Dai, Binyuan Hui, Min Yang, Zheng Cao, Jianbo Dong, Fei
  Huang, Luo Si, and Yongbin Li. 2022{\natexlab{a}}.
\newblock Space-2: Tree-structured semi-supervised contrastive pre-training for
  task-oriented dialog understanding.
\newblock In \emph{Proceedings of the 29th International Conference on
  Computational Linguistics}, pages 553--569.

\bibitem[{He et~al.(2022{\natexlab{b}})He, Dai, Yang, Sun, Huang, Si, and
  Li}]{he2022unified}
Wanwei He, Yinpei Dai, Min Yang, Jian Sun, Fei Huang, Luo Si, and Yongbin Li.
  2022{\natexlab{b}}.
\newblock Unified dialog model pre-training for task-oriented dialog
  understanding and generation.
\newblock In \emph{Proceedings of the 45th International ACM SIGIR Conference
  on Research and Development in Information Retrieval}, pages 187--200.

\bibitem[{He et~al.(2022{\natexlab{c}})He, Dai, Zheng, Wu, Cao, Liu, Jiang,
  Yang, Huang, Si, and Li}]{he2022galaxy}
Wanwei He, Yinpei Dai, Yinhe Zheng, Yuchuan Wu, Zheng Cao, Dermot Liu, Peng
  Jiang, Min Yang, Fei Huang, Luo Si, and Yongbin Li. 2022{\natexlab{c}}.
\newblock Galaxy: A generative pre-trained model for task-oriented dialog with
  semi-supervised learning and explicit policy injection.
\newblock In \emph{Proceedings of the AAAI Conference on Artificial
  Intelligence}, pages 10749--10757.

\bibitem[{Heck et~al.(2023)Heck, Lubis, Ruppik, Vukovic, Feng, Geishauser, Lin,
  van Niekerk, and Ga{\v{s}}i{\'c}}]{gasic}
Michael Heck, Nurul Lubis, Benjamin Ruppik, Renato Vukovic, Shutong Feng,
  Christian Geishauser, Hsien-Chin Lin, Carel van Niekerk, and Milica
  Ga{\v{s}}i{\'c}. 2023.
\newblock Chatgpt for zero-shot dialogue state tracking: A solution or an
  opportunity?
\newblock \emph{arXiv preprint arXiv:2306.01386}.

\bibitem[{Hu et~al.(2022{\natexlab{a}})Hu, Lin, Zhao, Lu, Wu, and
  Li}]{hu2022unimse}
Guimin Hu, Ting-En Lin, Yi~Zhao, Guangming Lu, Yuchuan Wu, and Yongbin Li.
  2022{\natexlab{a}}.
\newblock Unimse: Towards unified multimodal sentiment analysis and emotion
  recognition.
\newblock In \emph{Proceedings of the 2022 Conference on Empirical Methods in
  Natural Language Processing}, pages 7837--7851.

\bibitem[{Hu et~al.(2022{\natexlab{b}})Hu, Lee, Xie, Yu, Smith, and
  Ostendorf}]{text_sql}
Yushi Hu, Chia-Hsuan Lee, Tianbao Xie, Tao Yu, Noah~A Smith, and Mari
  Ostendorf. 2022{\natexlab{b}}.
\newblock In-context learning for few-shot dialogue state tracking.
\newblock \emph{arXiv preprint arXiv:2203.08568}.

\bibitem[{Hude{\v{c}}ek and Du{\v{s}}ek(2023)}]{24}
Vojt{\v{e}}ch Hude{\v{c}}ek and Ond{\v{r}}ej Du{\v{s}}ek. 2023.
\newblock Are llms all you need for task-oriented dialogue?
\newblock \emph{arXiv preprint arXiv:2304.06556}.

\bibitem[{Kojima et~al.(2022)Kojima, Gu, Reid, Matsuo, and
  Iwasawa}]{zero-shot_cot}
Takeshi Kojima, Shixiang~Shane Gu, Machel Reid, Yutaka Matsuo, and Yusuke
  Iwasawa. 2022.
\newblock Large language models are zero-shot reasoners.
\newblock \emph{Advances in neural information processing systems},
  35:22199--22213.

\bibitem[{Li et~al.(2022)Li, Lin, Zhang, Fu, Chen, Lou, and
  Chen}]{li2022advance}
Yifei Li, Zeqi Lin, Shizhuo Zhang, Qiang Fu, Bei Chen, Jian-Guang Lou, and
  Weizhu Chen. 2022.
\newblock On the advance of making language models better reasoners.
\newblock \emph{arXiv preprint arXiv:2206.02336}.

\bibitem[{Li et~al.(2023)Li, Lin, Wu, Liu, Tang, Zhao, and Li}]{li2023unisa}
Zaijing Li, Ting-En Lin, Yuchuan Wu, Meng Liu, Fengxiao Tang, Ming Zhao, and
  Yongbin Li. 2023.
\newblock Unisa: Unified generative framework for sentiment analysis.
\newblock \emph{arXiv preprint arXiv:2309.01339}.

\bibitem[{Lin et~al.(2022)Lin, Wu, Huang, Si, Sun, and Li}]{lin2022duplex}
Ting-En Lin, Yuchuan Wu, Fei Huang, Luo Si, Jian Sun, and Yongbin Li. 2022.
\newblock Duplex conversation: Towards human-like interaction in spoken
  dialogue systems.
\newblock In \emph{Proceedings of the 28th ACM SIGKDD Conference on Knowledge
  Discovery and Data Mining}, pages 3299--3308.

\bibitem[{Liu et~al.(2021)Liu, Shen, Zhang, Dolan, Carin, and
  Chen}]{liu2021makes}
Jiachang Liu, Dinghan Shen, Yizhe Zhang, Bill Dolan, Lawrence Carin, and Weizhu
  Chen. 2021.
\newblock What makes good in-context examples for gpt-$3 $?
\newblock \emph{arXiv preprint arXiv:2101.06804}.

\bibitem[{Lu et~al.(2021)Lu, Bartolo, Moore, Riedel, and
  Stenetorp}]{lu2021fantastically}
Yao Lu, Max Bartolo, Alastair Moore, Sebastian Riedel, and Pontus Stenetorp.
  2021.
\newblock Fantastically ordered prompts and where to find them: Overcoming
  few-shot prompt order sensitivity.
\newblock \emph{arXiv preprint arXiv:2104.08786}.

\bibitem[{Mannekote et~al.(2023)Mannekote, Celepkolu, Wiggins, and
  Boyer}]{schema_cate}
Amogh Mannekote, Mehmet Celepkolu, Joseph~B Wiggins, and Kristy~Elizabeth
  Boyer. 2023.
\newblock Exploring usability issues in instruction-based and schema-based
  authoring of task-oriented dialogue agents.
\newblock In \emph{Proceedings of the 5th International Conference on
  Conversational User Interfaces}, pages 1--6.

\bibitem[{Miao et~al.(2021)Miao, Liang, and Su}]{cot_arithmetic}
Shen-Yun Miao, Chao-Chun Liang, and Keh-Yih Su. 2021.
\newblock A diverse corpus for evaluating and developing english math word
  problem solvers.
\newblock \emph{arXiv preprint arXiv:2106.15772}.

\bibitem[{Mosig et~al.(2020)Mosig, Mehri, and Kober}]{star}
Johannes~EM Mosig, Shikib Mehri, and Thomas Kober. 2020.
\newblock Star: A schema-guided dialog dataset for transfer learning.
\newblock \emph{arXiv preprint arXiv:2010.11853}.

\bibitem[{Poria et~al.(2018)Poria, Hazarika, Majumder, Naik, Cambria, and
  Mihalcea}]{meld}
Soujanya Poria, Devamanyu Hazarika, Navonil Majumder, Gautam Naik, Erik
  Cambria, and Rada Mihalcea. 2018.
\newblock Meld: A multimodal multi-party dataset for emotion recognition in
  conversations.
\newblock \emph{arXiv preprint arXiv:1810.02508}.

\bibitem[{Qian et~al.(2023)Qian, Wang, Lin, Zheng, Zhu, Zhao, Hou, Wu, and
  Li}]{qian2023empathetic}
Yushan Qian, Bo~Wang, Ting-En Lin, Yinhe Zheng, Ying Zhu, Dongming Zhao,
  Yuexian Hou, Yuchuan Wu, and Yongbin Li. 2023.
\newblock Empathetic response generation via emotion cause transition graph.
\newblock \emph{arXiv:2302.11787}.

\bibitem[{Rastogi et~al.(2020)Rastogi, Zang, Sunkara, Gupta, and Khaitan}]{sgd}
Abhinav Rastogi, Xiaoxue Zang, Srinivas Sunkara, Raghav Gupta, and Pranav
  Khaitan. 2020.
\newblock Towards scalable multi-domain conversational agents: The
  schema-guided dialogue dataset.
\newblock In \emph{Proceedings of the AAAI conference on artificial
  intelligence}, volume~34, pages 8689--8696.

\bibitem[{Rubin et~al.(2021)Rubin, Herzig, and Berant}]{rubin2021learning}
Ohad Rubin, Jonathan Herzig, and Jonathan Berant. 2021.
\newblock Learning to retrieve prompts for in-context learning.
\newblock \emph{arXiv preprint arXiv:2112.08633}.

\bibitem[{Si et~al.(2023)Si, Ma, Wu, Dai, Gao, Lin, Li, Yan, Huang, and
  Li}]{spokenwoz}
Shuzheng Si, Wentao Ma, Yuchuan Wu, Yinpei Dai, Haoyu Gao, Ting-En Lin, Hangyu
  Li, Rui Yan, Fei Huang, and Yongbin Li. 2023.
\newblock Spokenwoz: A large-scale speech-text benchmark for spoken
  task-oriented dialogue in multiple domains.
\newblock \emph{arXiv preprint arXiv:2305.13040}.

\bibitem[{Sorensen et~al.(2022)Sorensen, Robinson, Rytting, Shaw, Rogers,
  Delorey, Khalil, Fulda, and Wingate}]{sorensen2022information}
Taylor Sorensen, Joshua Robinson, Christopher~Michael Rytting, Alexander~Glenn
  Shaw, Kyle~Jeffrey Rogers, Alexia~Pauline Delorey, Mahmoud Khalil, Nancy
  Fulda, and David Wingate. 2022.
\newblock An information-theoretic approach to prompt engineering without
  ground truth labels.
\newblock \emph{arXiv preprint arXiv:2203.11364}.

\bibitem[{Talmor et~al.(2018)Talmor, Herzig, Lourie, and
  Berant}]{cot_comonsense}
Alon Talmor, Jonathan Herzig, Nicholas Lourie, and Jonathan Berant. 2018.
\newblock Commonsenseqa: A question answering challenge targeting commonsense
  knowledge.
\newblock \emph{arXiv preprint arXiv:1811.00937}.

\bibitem[{Touvron et~al.(2023)Touvron, Martin, Stone, Albert, Almahairi,
  Babaei, Bashlykov, Batra, Bhargava, Bhosale et~al.}]{llama2}
Hugo Touvron, Louis Martin, Kevin Stone, Peter Albert, Amjad Almahairi, Yasmine
  Babaei, Nikolay Bashlykov, Soumya Batra, Prajjwal Bhargava, Shruti Bhosale,
  et~al. 2023.
\newblock Llama 2: Open foundation and fine-tuned chat models.
\newblock \emph{arXiv preprint arXiv:2307.09288}.

\bibitem[{Wang et~al.(2023)Wang, Xu, Lan, Hu, Lan, Lee, and
  Lim}]{plan-and-solve}
Lei Wang, Wanyu Xu, Yihuai Lan, Zhiqiang Hu, Yunshi Lan, Roy Ka-Wei Lee, and
  Ee-Peng Lim. 2023.
\newblock Plan-and-solve prompting: Improving zero-shot chain-of-thought
  reasoning by large language models.
\newblock \emph{arXiv preprint arXiv:2305.04091}.

\bibitem[{Wang et~al.(2022{\natexlab{a}})Wang, Wei, Schuurmans, Le, Chi,
  Narang, Chowdhery, and Zhou}]{self}
Xuezhi Wang, Jason Wei, Dale Schuurmans, Quoc Le, Ed~Chi, Sharan Narang,
  Aakanksha Chowdhery, and Denny Zhou. 2022{\natexlab{a}}.
\newblock Self-consistency improves chain of thought reasoning in language
  models.
\newblock \emph{arXiv preprint arXiv:2203.11171}.

\bibitem[{Wang et~al.(2022{\natexlab{b}})Wang, Wei, Schuurmans, Le, Chi, and
  Zhou}]{wang2022rationale}
Xuezhi Wang, Jason Wei, Dale Schuurmans, Quoc Le, Ed~Chi, and Denny Zhou.
  2022{\natexlab{b}}.
\newblock Rationale-augmented ensembles in language models.
\newblock \emph{arXiv preprint arXiv:2207.00747}.

\bibitem[{Wei et~al.(2022)Wei, Wang, Schuurmans, Bosma, Xia, Chi, Le, Zhou
  et~al.}]{cot}
Jason Wei, Xuezhi Wang, Dale Schuurmans, Maarten Bosma, Fei Xia, Ed~Chi, Quoc~V
  Le, Denny Zhou, et~al. 2022.
\newblock Chain-of-thought prompting elicits reasoning in large language
  models.
\newblock \emph{Advances in Neural Information Processing Systems},
  35:24824--24837.

\bibitem[{Yu et~al.(2023)Yu, Gao, Lin, Yang, Wu, Ma, Wang, Huang, and
  Li}]{yu2023speech}
Tianshu Yu, Haoyu Gao, Ting-En Lin, Min Yang, Yuchuan Wu, Wentao Ma, Chao Wang,
  Fei Huang, and Yongbin Li. 2023.
\newblock Speech-text pre-training for spoken dialog understanding with
  explicit cross-modal alignment.
\newblock In \emph{Proceedings of the 61st Annual Meeting of the Association
  for Computational Linguistics (Volume 1: Long Papers)}, pages 7900--7913.

\bibitem[{Zelikman et~al.(2022)Zelikman, Mu, Goodman, and
  Wu}]{zelikman2022star}
Eric Zelikman, Jesse Mu, Noah~D Goodman, and Yuhuai~Tony Wu. 2022.
\newblock Star: Self-taught reasoner bootstrapping reasoning with reasoning.

\bibitem[{Zhang et~al.(2022)Zhang, Zhang, Li, and Smola}]{zhang2022automatic}
Zhuosheng Zhang, Aston Zhang, Mu~Li, and Alex Smola. 2022.
\newblock Automatic chain of thought prompting in large language models.
\newblock \emph{arXiv preprint arXiv:2210.03493}.

\bibitem[{Zhao et~al.(2022)Zhao, Cao, Gupta, Lee, Rastogi, Wang, Soltau,
  Shafran, and Wu}]{starv2}
Jeffrey Zhao, Yuan Cao, Raghav Gupta, Harrison Lee, Abhinav Rastogi, Mingqiu
  Wang, Hagen Soltau, Izhak Shafran, and Yonghui Wu. 2022.
\newblock Anytod: A programmable task-oriented dialog system.
\newblock \emph{arXiv preprint arXiv:2212.09939}.

\bibitem[{Zhao et~al.(2021)Zhao, Wallace, Feng, Klein, and
  Singh}]{zhao2021calibrate}
Zihao Zhao, Eric Wallace, Shi Feng, Dan Klein, and Sameer Singh. 2021.
\newblock Calibrate before use: Improving few-shot performance of language
  models.
\newblock In \emph{International Conference on Machine Learning}, pages
  12697--12706. PMLR.

\end{thebibliography}
\bibliographystylelanguageresource{lrec-coling2024-natbib}
\bibliographylanguageresource{languageresource}

\end{document}